\documentclass[10pt,twocolumn,letterpaper]{article}

\usepackage{cvpr}              

\usepackage[dvipsnames]{xcolor}

\usepackage{amsmath,amsfonts,bm}

\def\eqref#1{equation~\ref{#1}}

\def\1{\bm{1}}

\DeclareMathAlphabet{\mathsfit}{\encodingdefault}{\sfdefault}{m}{sl}
\SetMathAlphabet{\mathsfit}{bold}{\encodingdefault}{\sfdefault}{bx}{n}

\definecolor{cvprblue}{rgb}{0.21,0.49,0.74}
\usepackage[pagebackref,breaklinks,colorlinks,citecolor=cvprblue]{hyperref}
\usepackage{url}
\usepackage{footmisc}
\usepackage{graphicx}
\usepackage{booktabs}
\usepackage{caption}
\usepackage{wrapfig}
\usepackage{adjustbox}
\usepackage[accsupp]{axessibility} 

\def\paperID{1769} 
\def\confName{CVPR}
\def\confYear{2024}

\begin{document}

\title{HumanGaussian: Text-Driven 3D Human Generation with Gaussian Splatting}

\author{Xian Liu$^{1}$, Xiaohang Zhan$^{2}$, Jiaxiang Tang$^{3}$, Ying Shan$^{2}$, Gang Zeng$^{3}$, Dahua Lin$^{1}$, Xihui Liu$^{4}$, Ziwei Liu$^{5}$ \\
$^1$CUHK \quad $^2$Tencent AI Lab \quad $^3$PKU \quad $^4$HKU \quad $^5$NTU \\
\href{https://alvinliu0.github.io/projects/HumanGaussian}{Project Page: https://alvinliu0.github.io/projects/HumanGaussian}
}

\twocolumn[{
\renewcommand\twocolumn[1][]{#1}
\maketitle
\pagestyle{empty}
\thispagestyle{empty}
\vspace{-30pt}
\begin{center}
\centering
\captionsetup{type=figure}
\includegraphics[width=\textwidth]{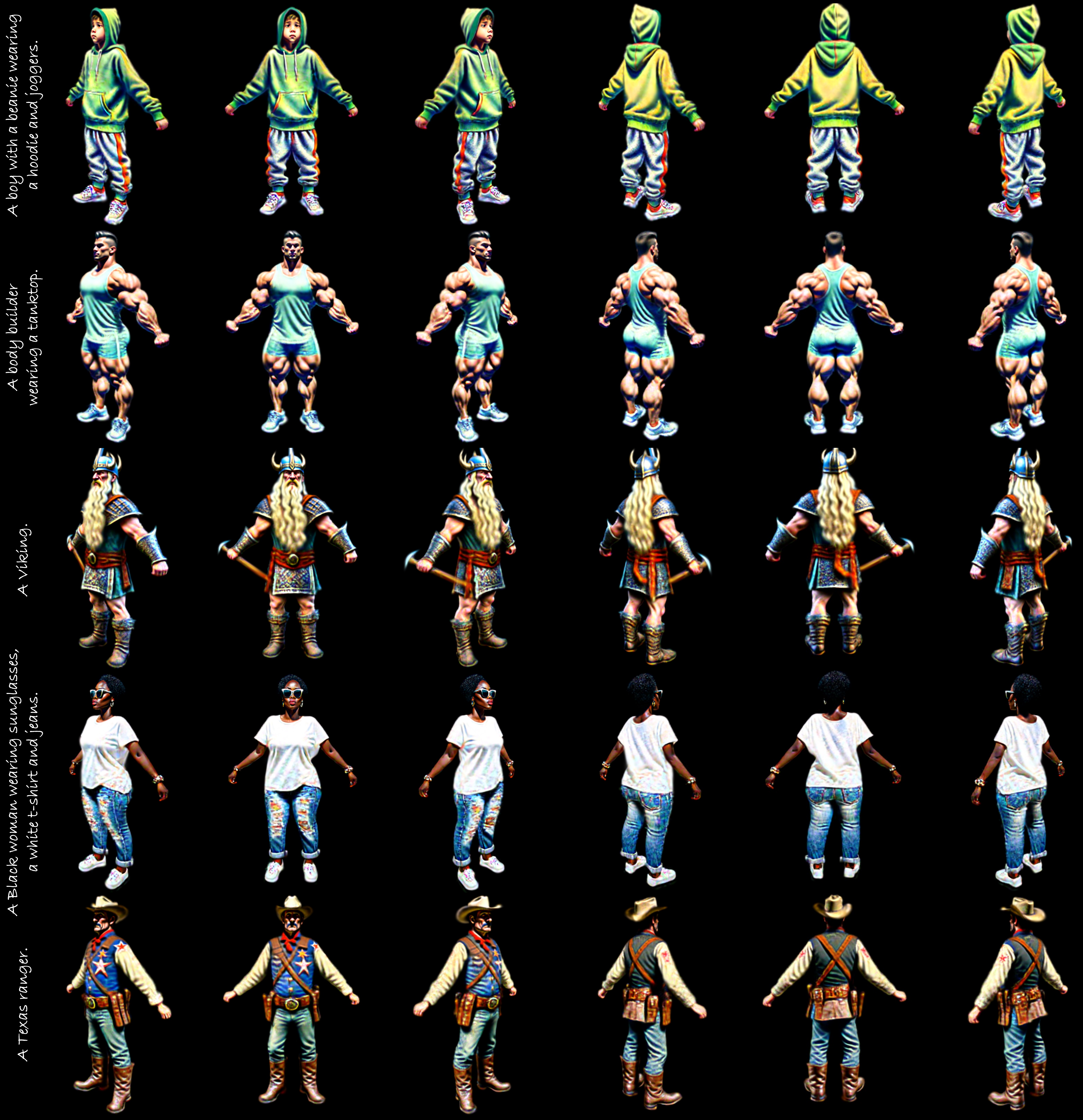}
\vspace{-20pt}
\captionof{figure}{
We propose \textbf{HumanGaussian}, an efficient yet effective framework that generates high-quality 3D humans with fine-grained geometry and realistic appearance. Our method adapts 3D Gaussian Splatting into text-driven 3D human generation with novel designs.
}
\label{fig:teaser}
\end{center}
}]

\maketitle
\pagestyle{empty}
\thispagestyle{empty}

\begin{abstract}

Realistic 3D human generation from text prompts is a desirable yet challenging task. 
Existing methods optimize 3D representations like mesh or neural fields via score distillation sampling (SDS), which suffers from inadequate fine details or excessive training time. 
In this paper, we propose an efficient yet effective framework, \textbf{HumanGaussian}, that generates high-quality 3D humans with fine-grained geometry and realistic appearance. 
Our key insight is that 3D Gaussian Splatting is an efficient renderer with periodic Gaussian shrinkage or growing, where such adaptive density control can be naturally guided by intrinsic human structures. 
Specifically, \textbf{1)} we first propose a Structure-Aware SDS that simultaneously optimizes human appearance and geometry. The multi-modal score function from both RGB and depth space is leveraged to distill the Gaussian densification and pruning process. 
\textbf{2)} Moreover, we devise an Annealed Negative Prompt Guidance by decomposing SDS into a noisier generative score and a cleaner classifier score, which well addresses the over-saturation issue. The floating artifacts are further eliminated based on Gaussian size in a prune-only phase to enhance generation smoothness. 
Extensive experiments demonstrate the superior efficiency and competitive quality of our framework, rendering vivid 3D humans under diverse scenarios.

\end{abstract} 
\section{Introduction}
\label{sec:intro}

Creating high-quality 3D humans from user condition is of great importance to a wide variety of applications, ranging from virtual try-on~\cite{wang2018toward, santesteban2021self, santesteban2022ulnef, ju2023humansd} to immersive telepresence~\cite{li2020volumetric, saito2020pifuhd, zhu2023taming, liu2022audio, ju2023direct, ju2023human}. To this end, researchers explore the task of text-driven 3D human generation, which synthesizes the character's appearance and geometry based on text prompts. Traditional methods resort to a hand-crafted pipeline, where 3D models are first regressed from multi-view human captures, and then undergo a series of manual processes like rigging and skinning~\cite{bickel2007multi, joo2015panoptic, li2021hybrik, kocabas2021pare}. To ease human labor for 3D asset creation of diverse layouts, the exemplar work DreamFusion~\cite{poole2022dreamfusion} proposes score distillation sampling (SDS) to harness rich 2D text-to-image prior (\textit{e.g.}, Stable Diffusion~\cite{rombach2022high}, Imagen~\cite{saharia2022photorealistic}) by optimizing 3D scenes to render samples that reside on the manifold of higher likelihood. Though accomplishing reasonable results on single objects~\cite{richardson2023texture, metzer2023latent, wang2023prolificdreamer, chen2023fantasia3d}, it is hard for them to model detailed human bodies with complex articulations. 

To incorporate structural guidance, recent text-driven 3D human studies combine SDS with body shape models such as SMPL~\cite{loper2023smpl} and imGHUM~\cite{alldieck2021imghum}. In particular, a common paradigm is to integrate human priors into representations like mesh and neural radiance field (NeRF), either by taking the body shape as mesh/density initialization~\cite{hong2022avatarclip, zeng2023avatarbooth, kolotouros2023dreamhuman}, or by learning a deformation field based on linear blend skinning (LBS)~\cite{cao2023dreamavatar, weng2022humannerf, yu2023monohuman}. However, they mostly compromise to trade-off between efficiency and quality: the mesh-based methods~\cite{liao2023tada, jiang2023avatarcraft, youwang2022clip} struggle to model fine topologies like accessories and wrinkles; while the NeRF-based methods~\cite{huang2023humannorm, zhang2023avatarverse, huang2023dreamwaltz} are time/memory-consuming to render high-resolution results. How to achieve fine-grained generation efficiently remains an unsolved problem.

Recently, the explicit neural representation of 3D Gaussian Splatting (3DGS)~\cite{kerbl20233d} provides a new perspective for real-time scene reconstruction. It enables multi-scale modeling across multiple granularities, which is suitable for 3D human generation. Nevertheless, it is non-trivial to exploit such representation in this task with two challenges: 
\textbf{1)} 3DGS characterizes a tile-based rasterization by sorting and $\alpha$-blending anisotropic splats within each view frustum, which only back-propagates a small set of high-confidence Gaussians. However, as verified in the 3D surface-/volume-rendering studies~\cite{hart1996sphere, takikawa2021neural, kajiya1984ray, mildenhall2021nerf}, sparse gradient could hinder network optimization of geometry and appearance. Therefore, structural guidance is required in 3DGS, especially for the human domain that demands hierarchical structure modeling and generation controllability. 
\textbf{2)} The naive SDS necessitates a large classifier-free guidance (CFG)~\cite{ho2022classifier} scale for image-text alignment (\textit{e.g.}, $100$ as used in~\cite{poole2022dreamfusion}). But it sacrifices visual quality with over-saturated patterns, making realistic human generation difficult. Besides, due to the stochasticity of SDS loss, the original gradient-based density control in 3DGS is unstable, which incurs blurry results with floating artifacts.

In this paper, we propose an efficient yet effective framework, \textbf{HumanGaussian}, that generates high-quality 3D humans with fine-grained geometry and realistic appearance. Our intuition lies in that 3D Gaussian Splatting is an efficient renderer with periodic Gaussian shrinkage or growing, where \textit{such adaptive density control can be naturally guided by intrinsic human structures}. The key is to incorporate explicit structural guidance and gradient regularization to facilitate Gaussian optimization.
Specifically, we first propose a \textit{Structure-Aware SDS} that jointly learns human appearance and geometry. Unlike previous studies~\cite{yi2023gaussiandreamer, chen2023text, kerbl20233d} that adopt the generic priors like Structure-from-Motion (SfM) points or Point-E~\cite{nichol2022point}, we instead anchor the Gaussian initial positions on the SMPL-X mesh. The subsequent densification and pruning processes thus focus on regions around the body surface, effectively capturing geometric deformations like accessories and wrinkles. Additionally, we extend the pre-trained Stable Diffusion~\cite{rombach2022high} to simultaneously denoise the image RGB and depth as our SDS source model. Such dual-branch design distills the joint distribution of two spatially-aligned targets (\textit{i.e.}, RGB and depth), which boosts the Gaussian convergence with both structural guidance and textural realism. 
To further improve renderings with natural appearance, we devise an \textit{Annealed Negative Prompt Guidance}. In particular, we decompose SDS into a noisier generative score and a cleaner classifier score, where equipping the latter term with a decreasing negative prompt guidance enables realistic generation under nominal CFG scales (\textit{e.g.}, $7.5$), as also proven in concurrent text-to-3D studies~\cite{katzir2023noise, yu2023text}. In this way, we manage to avoid over-saturated patterns with appropriate CFG scales that well balance sample quality and diversity. Moreover, due to the high variance of SDS loss, directly relying on gradient information to control densities results in blurry geometry~\cite{kerbl20233d}. In contrast, we propose to eliminate floating artifacts based on Gaussian size in a prune-only phase. 

To summarize, our main contributions are three-fold: 
\textbf{1)} We propose an efficient yet effective framework \textbf{HumanGaussian} for high-quality 3D human generation with fine-grained geometry and realistic appearance. As one of the earliest attempts in taming Gaussian Splatting for text-driven 3D human domain, we hope to pave the way for future research. 
\textbf{2)} We propose the \textit{Structure-Aware SDS} to jointly learn human appearance and geometry with explicit structural guidance. 
\textbf{3)} We devise the \textit{Annealed Negative Prompt Guidance} to guarantee realistic results and eliminate floating artifacts. Extensive experiments demonstrate the superior efficiency and competitive quality of our framework, rendering vivid 3D humans under diverse scenarios.
\section{Related Work}
\label{sec:related}
\noindent\textbf{3D Neural Representations.} 
Diverse 3D scene representations are proposed for spatial geometry and texture modeling, such as voxel, point cloud, mesh, and neural field. With the trade-off among training time, memory efficiency, rendering capability, and network compatibility, different representations are chosen based on problem setting: \textbf{1)} Voxel, a Euclidean representation that stores scene information in a grid manner~\cite{wu20153d, maturana2015voxnet, choy20163d}, can be easily adapted for CNNs, but is limited in render resolution due to the cubic computational cost. \textbf{2)} Point cloud, a discrete point set sampled from 3D surface, is efficient to render~\cite{qi2017pointnet, qi2017pointnet++, li2018pointcnn}. However, it fails to capture the fine-grained details due to its discontinuous nature. \textbf{3)} Mesh, a compact representation expressing the connectivity among vertices, edges, and faces, inherits time efficiency from the well-rounded graphic pipelines~\cite{wang2018pixel2mesh, wen2019pixel2mesh++, hanocka2019meshcnn}, but struggles to create accurate topology. \textbf{4)} Neural field, an implicit function of each 3D position's attributes, is capable of modeling complex structures in arbitrary resolution~\cite{oechsle2021unisurf, mildenhall2021nerf, yariv2021volume, liu2022semantic, wu2022object}, yet the optimization and inference are slow. Recently, 3D Gaussian Splatting (3DGS)~\cite{kerbl20233d, luiten2023dynamic} has shown impressive results in the 3D reconstruction, surpassing previous representations with better quality and faster convergence. In this work, we try to unlock the potential of 3D Gaussian splatting on the challenging task of text-driven 3D human generation.

\begin{figure*}[t]
    \centering
    \includegraphics[width=\linewidth]{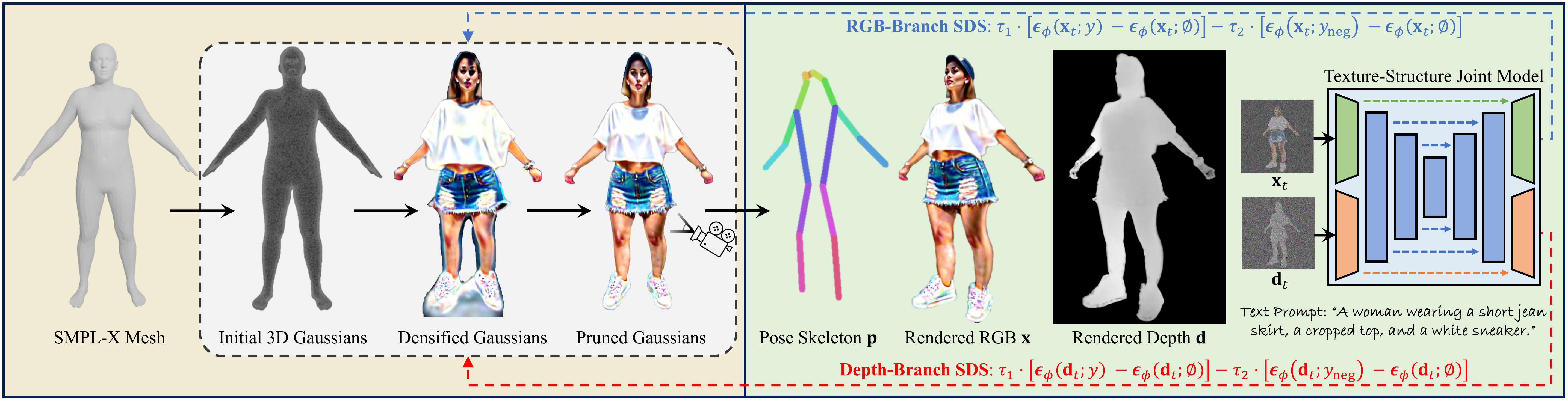}
    \vspace{-17pt}
    \caption{\textbf{Overview of the proposed \textbf{HumanGaussian} Framework.} We generate high-quality 3D humans from text prompts with the neural representation of 3D Gaussian Splatting (3DGS). In \textit{Structure-Aware SDS}, we start from the SMPL-X prior to densely sample Gaussians on the human mesh surface as initial center positions. Then, a Texture-Structure Joint Model is trained to simultaneously denoise the image $\mathbf{x}$ and depth $\mathbf{d}$ conditioned on pose skeleton $\mathbf{p}$. Based on this, we design a dual-branch SDS to jointly optimize human appearance and geometry, where the 3DGS density is adaptively controlled by distilling from both the RGB and depth space. In \textit{Annealed Negative Prompt Guidance}, we use the cleaner classifier score with an annealed negative score to regularize the stochastic SDS gradient of high variance. The floating artifacts are further eliminated based on Gaussian size in a prune-only phase to enhance generation smoothness.}
    \vspace{-10pt}
    \label{fig:pipeline}
\end{figure*}

\noindent\textbf{Text-to-3D Generation.} 
Recent diffusion-based text-to-3D works can be grouped into two types: \textbf{1)} 3D native pipelines, which directly capture the distribution of 3D data~\cite{jun2023shap, nichol2022point, lorraine2023att3d} or the reconstructed intermediate features~\cite{cheng2023sdfusion, chen2023single, gupta20233dgen, ntavelis2023autodecoding} on specific domains. Although some recent works~\cite{anonymous2023instantd, hong2023lrm} extend the model's capacity by training on large-scale 3D datasets like Objaverse~\cite{deitke2023objaverse}, they are still confined to single objects. \textbf{2)} Optimization-based 2D lifting pipelines, which optimize 3D scene representations in a per-prompt manner by distilling from rich prior learned in the 2D domain. For example, some early attempts use CLIP guidance~\cite{radford2021learning} to boost multi-view image-text alignment~\cite{jain2022zero, wang2022clip, mohammad2022clip, michel2022text2mesh}, while recent methods resort to score distillation sampling (SDS) to inherit unprecedented rendering quality from exemplar text-to-image models~\cite{poole2022dreamfusion, wang2023score, metzer2023latent, wang2023prolificdreamer, chen2023fantasia3d}. Notably, the heavy computation burden of NeRF incurs long training time, which motivates concurrent works to adapt the representation of Gaussian splatting for text-to-3D generation~\cite{tang2023dreamgaussian, chen2023text, yi2023gaussiandreamer}. In this work, we choose 3D Gaussian due to its efficiency and efficacy, but focus on the text-driven 3D human domain that demands both fine detail capturing and realistic texture generation.

\noindent\textbf{Text-Driven 3D Human Generation.} 
By incorporating human prior like SMPL~\cite{loper2023smpl} and imGHUM~\cite{alldieck2021imghum} models, the 3D text-to-human literature typically adapts general text-to-3D techniques with domain-specific designs~\cite{hong2022eva3d, cao2023dreamavatar, jiang2023avatarcraft, zeng2023avatarbooth, zhang2023text, gao2023textdeformer, zhao2023zero, liu2022learning}. For example, AvatarCLIP~\cite{hong2022avatarclip} integrates NeuS~\cite{wang2021neus} with SMPL prior and utilizes CLIP for guidance. DreamHuman~\cite{kolotouros2023dreamhuman} utilizes a pose-conditioned NeRF based on imGHUM to learn albedo and density fields with SDS. AvatarVerse~\cite{zhang2023avatarverse} fine-tunes the ControlNet~\cite{zhang2023adding} branch with DensePose~\cite{guler2018densepose} as SDS source for view-consistent generation. TADA~\cite{liao2023tada} deforms the SMPL-X~\cite{pavlakos2019expressive} shape with displacement and optimizes texture UV-map by hierarchical rendering. Concurrent to our work, HumanNorm~\cite{huang2023humannorm} also adds explicit structural constraint with the fine-tuned text-to-depth/normal models. However, they characterize a two-stage pipeline with DMTet~\cite{shen2021deep} representation, which fails to capture the fine-grained human details efficiently.
\section{Our Approach}

We present \textbf{HumanGaussian} that generates high-quality 3D humans with fine-grained geometry and realistic appearance. The overall framework is illustrated in Fig.~\ref{fig:pipeline}. To make the content self-contained and narration clearer, we first introduce some pre-requisites in Sec.~\ref{sec:3.1}. Then, we present the \textit{Structure-Aware SDS} which jointly learns human appearance and geometry with explicit structural guidance. The multi-modal score function from both RGB and depth space is leveraged to distill the Gaussian densification and pruning process (Sec.~\ref{sec:3.2}). Finally, we elaborate the \textit{Annealed Negative Prompt Guidance} to guarantee realistic results and eliminate floating artifacts in Sec.~\ref{sec:3.3}.

\subsection{Preliminaries}
\label{sec:3.1}

\noindent\textbf{SMPL-X}~\cite{pavlakos2019expressive} is a 3D parametric human model that defines the shape topology of body, hands, and face. It contains $10,475$ vertices and $54$ keypoints. By assembling pose parameters $\theta$ (consists of body pose $\theta_b$, jaw pose $\theta_f$, and finger pose $\theta_h$), shape parameters $\beta$, and expression parameters $\psi$, we can represent 3D SMPL-X human model $M(\beta, \theta, \psi)$ as:
\begin{align}
\label{eq:smplx}
\begin{aligned}
    T(\beta, \theta, \psi) &= \bar{T} + B_s(\beta) + B_p(\theta) + B_e(\psi), \\
    M(\beta, \theta, \psi) &= \mathtt{LBS}(T(\beta, \theta, \psi), J(\beta), \theta, \mathcal{W}),
\end{aligned}
\end{align}
where $\bar{T}$ is the mean template shape; $B_s$, $B_p$, and $B_e$ are the blend shape functions for shape, pose, and expression, respectively; $T(\beta, \theta, \psi)$ is the non-rigid deformation from $\bar{T}$; $\mathtt{LBS}(\cdot)$ is the linear blend skinning function~\cite{lewis2023pose} that transforms $T(\beta, \theta, \psi)$ into target pose $\theta$, with the skeleton joints $J(\beta)$ and blend weights $\mathcal{W}$ defined on each vertex. 

\noindent\textbf{Score Distillation Sampling (SDS)} is proposed in DreamFusion~\cite{poole2022dreamfusion} to distill the 2D pre-trained diffusion prior for optimizing 3D representations. Specifically, we represent a 3D scene parameterized by $\theta$ and use a  differentiable rendering function $g(\cdot)$ to obtain an image $\mathbf{x} = g(\theta)$. By pushing samples towards denser regions of the real-data distribution across all noise levels, we make renderings from each camera view resemble the plausible samples derived from the guidance diffusion model $\phi$. In practice, DreamFusion uses Imagen~\cite{saharia2022photorealistic} as the score estimation function $\bm{\epsilon}_{\phi}(\mathbf{x}_t; y)$, which predicts the sampled noise $\bm{\epsilon}_{\phi}$ given the noisy image $\mathbf{x}_t$, text embedding $y$, and timestep $t$. SDS optimizes 3D scenes using gradient descent with respect to $\theta$:
\begin{align}
\label{eq:sds}
\nabla_\theta \mathcal{L}_{\mathrm{SDS}}=\mathbb{E}_{\bm{\epsilon}, t}\left[w_t\left(\bm{\epsilon}_\phi\left(\mathbf{x}_t ; y\right)-\bm{\epsilon}\right) \frac{\partial \mathbf{x}}{\partial \theta}\right],
\end{align}
where $\bm{\epsilon}\sim\mathcal{N}(\mathbf{0}, \mathbf{I})$ is a Gaussian noise; $\mathbf{x}_t=\alpha_t\mathbf{x}+\sigma_t\bm{\epsilon}$ is the noised image; $\alpha_t$, $\sigma_t$, and $w_t$ are noise sampler terms.

\noindent\textbf{3D Gaussian Splatting}~\cite{kerbl20233d} features an effective representation for novel-view synthesis and 3D reconstruction. Different from those implicit counterparts like NeRF~\cite{mildenhall2021nerf}, 3D Gaussian Splatting represents the underlying scene through a set of anisotropic Gaussians parameterized by their center position $\mu \in \mathbb{R}^3$, covariance $\Sigma \in \mathbb{R}^7$, color $c \in \mathbb{R}^3$, and opacity $\alpha \in \mathbb{R}$. By projecting 3D Gaussians onto the camera's imaging plane, the 2D Gaussians are assigned to the corresponding tiles for point-based rendering~\cite{zwicker2001ewa}:
\begin{align}
\label{eq:gs}
\begin{aligned}
    G\left(p, \mu_i, \Sigma_i\right) = \exp (-\frac{1}{2} (p - \mu_i)^\mathsf{T} \Sigma_i^{-1} (p - \mu_i))&, \\
    \mathbf{c}(p)=\sum_{i \in \mathcal{N}} c_i \sigma_i \prod_{j=1}^{i-1}\left(1-\sigma_j\right), \sigma_i=\alpha_i G\left(p, \mu_i, \Sigma_i\right)&,
\end{aligned}
\end{align}
where $p$ is the location of queried point; $\mu_i$, $\Sigma_i$, $c_i$, $\alpha_i$, and $\sigma_i$ are the center position, covariance, color, opacity, and density of the $i$-th Gaussian, respectively; $G\left(p, \mu_i, \Sigma_i\right)$ is the value of the $i$-th Gaussian at point $p$; $\mathcal{N}$ denotes the set of 3D Gaussians in this tile. Besides, 3D Gaussian Splatting improves a GPU-friendly rasterization process with better quality, faster rendering speed, and less memory usage.

\subsection{Structure-Aware SDS}
\label{sec:3.2}
To adapt Gaussian Splatting for text-driven 3D human generation, the simplest way is by substituting scene representations of mesh or implicit neural fields with explicit 3DGS~\cite{kerbl20233d}. However, three problems remain: 
\textbf{1)} The original 3DGS training process heavily relies on center position initialization, even for the comparatively easier 3D reconstruction with dense multi-view image supervision. How to initialize Gaussians in the text-to-3D generation setting, especially for the highly-detailed human domain. 
\textbf{2)} To ease 3DGS learning with both appearance and geometry guidance, we have to supplement structural knowledge like text-to-depth/normal to text-to-image. While previous studies mostly capture a singular modality, how to simultaneously learn the joint distribution of structure and texture remains challenging. 
\textbf{3)} The commonly-used SDS~\cite{poole2022dreamfusion} distills solely from RGB space, yet it is hard to optimize the point-based 3DGS renderings with sparse gradients. How to add explicit structural supervision for better convergence. 

\noindent\textbf{Gaussian Initialization with SMPL-X Prior.} 
Our solution to the first problem is to initialize 3D Gaussian center positions based on the SMPL-X mesh shape. We choose it as human domain-specific structural prior due to two reasons: \textbf{1)} Previous studies~\cite{kerbl20233d, yi2023gaussiandreamer, chen2023text} use either Structure-from-Motion (SfM) points~\cite{snavely2006photo} or generic text-to-point-cloud priors of Shap-E~\cite{jun2023shap} and Point-E~\cite{nichol2022point}. However, such methods typically fall short in the human category, resulting in over-sparse points or incoherent body structures. \textbf{2)} As an extension to SMPL~\cite{loper2023smpl}, SMPL-X complements the shape topology of face and hands, which are beneficial to intricate human modeling with fine-grained details. Based on such observations, we propose to uniformly sample points over SMPL-X mesh surface as 3DGS initialization. Specifically, $100k$ 3D Gaussians are instantiated with unit scaling, mean color, and no rotation, which are significantly denser than SMPL-X-defined vertices to model local details. We scale and transform the 3DGS to make it reasonable human size and located in the center of 3D space. To further extract 2D skeleton as structural condition, we map SMPL-X joints to the COCO-style human keypoints. When drawing keypoints on canvas as body skeleton maps, we carefully cull the occluded joints for more accurate visualization, such as removing the left/right eye and ear keypoints from the right-/left-side view, and hiding face keypoints from back view. 

\noindent\textbf{Learn Texture-Structure Joint Distribution.} 
Since the SMPL-X prior merely serves as an initialization, more comprehensive guidance is needed to facilitate 3DGS training. Instead of distilling the 3D scene from a single-modality diffusion model that solely learns the appearance~\cite{saharia2022photorealistic} or geometry~\cite{gupta20233dgen}, we propose to harvest an SDS source model capturing the joint distribution of both texture and structure. In particular, we take inspiration from a recent work~\cite{liu2023hyperhuman} to extend the pre-trained Stable Diffusion~\cite{rombach2022high} with structural expert branches to denoise the image RGB and depth simultaneously. By trading off between the spatial alignment and accurate distribution learning, we replicate the diffusion UNet backbone's $\mathtt{conv\_in}$, the first $\mathtt{DownBlock}$, the last $\mathtt{UpBlock}$, and $\mathtt{conv\_out}$ layers to deal with the denoising of each target. To further enable flexible skeleton control, we additionally take pose map $\mathbf{p}$ as an input condition via channel-wise concatenation. The overall network is trained with $\mathbf{v}$-prediction~\citep{salimans2022progressive} to minimize the simplified objective:
\begin{align} \label{eq:tp2rd}
\small
    \mathbb{E}_{\bm{\mathbf{v}}, t} \left[\lvert\lvert\bm{\mathbf{v}}_\phi(\mathbf{x}_t; \mathbf{p}, y) - \bm{\mathbf{v}}_t^\mathbf{x} \rvert\rvert_2^2 + 
    \lvert\lvert\bm{\mathbf{v}}_\phi(\mathbf{d}_t; \mathbf{p}, y) -\bm{\mathbf{v}}_t^{\mathbf{d}} \rvert\rvert_2^2\right],
\end{align} 
where $\mathbf{x}$ and $\mathbf{d}$ are the image-depth pairs annotated from large dataset; $\mathbf{x}_t=\alpha_t\mathbf{x}+\sigma_t\bm{\epsilon}_{\mathbf{x}}$ and $\mathbf{d}_t=\alpha_t\mathbf{d}+\sigma_t\bm{\epsilon}_{\mathbf{d}}$ are the noised features of RGB and depth; 
$\bm{\epsilon}_{\mathbf{x}}, \bm{\epsilon}_{\mathbf{d}}\sim\mathcal{N}(\mathbf{0}, \mathbf{I})$ are independently sampled noise; 
$\bm{\mathbf{v}}_t^{\mathbf{x}}=\alpha_t \bm{\epsilon}_{\mathbf{x}}-\sigma_t \mathbf{x}$ and $\bm{\mathbf{v}}_t^{\mathbf{d}}=\alpha_t \bm{\epsilon}_{\mathbf{d}}-\sigma_t \mathbf{d}$ are the $\mathbf{v}$-prediction learning targets at time-step $t$ for the RGB and depth, respectively. 
In this way, we obtain a unified model that captures both the image texture of appearance and structure of fore-/back-ground relationship, which can be used in SDS to boost 3DGS learning.

\noindent\textbf{Dual-Branch SDS as Optimization Guidance.} 
With the extended diffusion model that produces the spatially aligned image RGB and depth, we can guide the 3DGS optimization process from both structural and textural aspects. Specifically, the per-view depth map at each pixel can be computed by accumulating depth value overlapping the pixel of $\mathcal{N}$ ordered Gaussian instances via point-based $\alpha$-blending: 
\begin{align}
\label{eq:gs-depth}
    \mathbf{d}(p)=\sum_{i \in \mathcal{N}} d_i\sigma_i \prod_{j=1}^{i-1}\left(1-\sigma_j\right), \sigma_i=\alpha_i G\left(p, \mu_i, \Sigma_i\right)&,
\end{align}
where $d_i$ is the projected depth of the $i$-th Gaussian center $\mu_i$ in the current camera view; $G\left(p, \mu_i, \Sigma_i\right)$ is the value of the $i$-th Gaussian at the queried point $p$ as defined in Eq.~\ref{eq:gs}. Afterward, all the rendered depth maps $\mathbf{d}$ are normalized to the data range of $[0, 1]$. Combined with image renderings $\mathbf{x}$, we can optimize 3DGS with an improved dual-branch SDS:
\begin{align}
\label{eq:dual-sds}
\begin{aligned}
\nabla_\theta \mathcal{L}_{\mathrm{SDS}} &= \lambda_1 \cdot \mathbb{E}_{\bm{\epsilon}_{\mathbf{x}}, t}\left[w_t\left(\bm{\epsilon}_\phi\left(\mathbf{x}_t ; \mathbf{p}, y\right)-\bm{\epsilon}_{\mathbf{x}}\right) \frac{\partial \mathbf{x}}{\partial \theta}\right] \\
&+ \lambda_2 \cdot \mathbb{E}_{\bm{\epsilon}_{\mathbf{d}}, t}\left[w_t\left(\bm{\epsilon}_\phi\left(\mathbf{d}_t ; \mathbf{p}, y\right)-\bm{\epsilon}_{\mathbf{d}}\right) \frac{\partial \mathbf{d}}{\partial \theta}\right],\\
\end{aligned}
\end{align}
where $\lambda_1$ and $\lambda_2$ are coefficients balancing the effects from the structural and textural branches; $\bm{\epsilon}_\phi(\cdot)$ are the $\bm{\epsilon}$-predictions derived from joint texture-structure model in Eq.~\ref{eq:tp2rd}. Such structural regularization helps reduce geometry distortions, thus benefiting the optimization of 3DGS with sparse gradient information. Note that although a concurrent work~\cite{huang2023humannorm} also adds structural constraints by finetuning two \textit{individual} text-to-normal and text-to-depth models for SDS, they suffer from misaligned depth and normal predictions, which could potentially mislead the network training. 

\subsection{Annealed Negative Prompt Guidance}
\label{sec:3.3}
\noindent\textbf{Decompose SDS with Classifier-Free Guidance.} 
To enforce text-3D alignment of high quality, DreamFusion~\cite{poole2022dreamfusion} uses a large classifier-free guidance scale to update the score matching difference term $\delta$ for 3D scene optimization:
\begin{align}
\label{eq:sds-cfg}
\delta=\underbrace{\left[\bm{\epsilon}_\phi\left(\mathbf{x}_t ; y\right)-\bm{\epsilon}\right]}_{\textit{generative score }\delta_g} \; + \; \tau \cdot \underbrace{\left[\bm{\epsilon}_\phi\left(\mathbf{x}_t ; y\right)-\bm{\epsilon}_\phi\left(\mathbf{x}_t ; \varnothing\right)\right]}_{\textit{classifier score }\delta_c},
\end{align}
where $\tau$ is the classifier-free guidance scale; $\varnothing$ denotes the null condition of an empty prompt. In this formulation, we can naturally decompose score matching difference into two parts $\delta=\delta_g + \tau \cdot \delta_c$, where the former term is a generative score that pushes images towards more realistic manifold; and the latter term is a classifier score that aligns samples with an implicit classifier~\cite{katzir2023noise, yu2023text}. However, as the generative score contains Gaussian noise $\bm{\epsilon}$ of high variance, it provides stochastic gradient information that harms the training stability. To deal with this, Poole \textit{et al.}~\cite{poole2022dreamfusion} deliberately intensify CFG scale $\tau$ to make classifier score dominate the optimization, leading to over-saturated patterns. Instead, we only leverage the cleaner classifier score $\delta_c$ as SDS loss.

\noindent\textbf{Regularize Gradient with Negative Prompts.} 
The negative prompts are widely used in text-to-image to bypass undesired properties. In view of this, we propose to augment classifier score for better 3DGS learning. Specifically, we substitute the null condition $\varnothing$ with negative prompts $y_{\text{neg}}$ to derive the improved negative classifier score $\delta_{nc}$:
\begin{align}
\label{eq:neg-cls}
\begin{aligned}
\delta_{nc}&=\bm{\epsilon}_\phi\left(\mathbf{x}_t ; y\right)-\bm{\epsilon}_\phi\left(\mathbf{x}_t ; y_{\text{neg}}\right), \\
&=\underbrace{\left[\bm{\epsilon}_\phi\left(\mathbf{x}_t ; y\right)-\bm{\epsilon}_\varnothing\right]}_{\textit{classifier score }\delta_c} - \underbrace{\left[\bm{\epsilon}_\phi\left(\mathbf{x}_t ; y_{\text{neg}}\right)-\bm{\epsilon}_\varnothing\right]}_{\textit{negative score }\delta_n},
\end{aligned}
\end{align}
where $\bm{\epsilon}_\varnothing = \bm{\epsilon}_\phi\left(\mathbf{x}_t ; \varnothing\right)$ is the abbreviation for formula conciseness. In this way, we regularize SDS gradients from two aspects: the rendered image is not only guided to align with the implicit classifier, but also forced to repel from the negative mode. Empirically, we find the negative score harm quality at small timesteps~\cite{katzir2023noise, yu2023text}. So we use an annealed negative guidance to combine both scores for supervision:
\begin{align}
\label{eq:neg-sds}
\nabla_\theta \mathcal{L}_{\mathrm{SDS}}=\mathbb{E}_{\bm{\epsilon}, t}\left[w_t\left(\tau_1 \cdot \delta_c - \tau_2 \cdot \delta_n\right) \frac{\partial \mathbf{x}}{\partial \theta}\right],
\end{align}
where $\tau_1$ and $\tau_2$ balance the effects from classifier and negative scores, with $\tau_2$ gradually drops at smaller timesteps. Note that such guidance can also be applied to depth. In practice, we use Eq.~\ref{eq:neg-sds} for both image and depth branches.

\noindent\textbf{Size-Conditioned Gaussian Pruning.} 
Although the proposed annealed negative prompt guidance provides cleaner gradients for 3DGS training, the SDS-based supervision is still much more stochastic than reconstruction-based error, leading to vaporous blurriness near the human surface. One observation is that such floating artifacts emerge as large-size tiny-opacity Gaussians during the densification process of cloning and splitting. To this end, we propose to eliminate them based on the scaling factor in a prune-only phase (covariance $\Sigma \in \mathbb{R}^7$ can be decoupled as a scaling factor $s \in \mathbb{R}^3$ and a rotation quaternion $q \in \mathbb{R}^4$). Particularly, after the adaptive density control that heavily relies on high-variance SDS, we extend a prune-only phase to remove Gaussian instances whose scaling factor is above a certain threshold. Note that a potential problem is that some useful Gaussians could be mistakenly eliminated. Despite this, in practice we find such a mechanism robust to maintain fine-grained geometry for two reasons: \textbf{1)} Those tiny-opacity Gaussians contribute little to point-based $\alpha$-blending, which are negligible in the rendered results. \textbf{2)} Thanks to the dense initialization based on SMPL-X prior, the Gaussians are redundant near the human body surface, which could compensate for the detail loss during optimization.
\section{Experiments}
\label{sec:exp}

\subsection{Implementation Details}
\noindent\textbf{SDS Guidance Model Setups.} 
As elaborated in Sec.~\ref{sec:3.2}, we extend the pretrained SD to capture the joint distribution of texture and structure by simultaneously denoising RGB and depth. The depth maps are labeled by MiDaS~\citep{Ranftl2022} on LAION~\citep{schuhmann2022laion}. The model is finetuned from \textit{SD 2.0} with $\mathbf{v}$-prediction~\cite{salimans2022progressive} in $512$ resolution. The DDIM scheduler~\cite{song2020denoising} is used with classifier-score weight $\tau_1$ as $7.5$. We gradually drop the negative-score weight $\tau_2$ from $1.0$ to $0.0$ at timestep $200$ for annealed negative prompt SDS guidance.

\noindent\textbf{3D Gaussian Splatting Setups.} 
The 3D Gaussians are initialized with $100k$ instances evenly sampled on SMPL-X mesh surface with opacity of $0.1$. The color is represented by Spherical Harmonics (SH) coefficients~\cite{fridovich2022plenoxels} of degree $0$ following~\cite{kerbl20233d}. The whole 3DGS training takes $3600$ iterations, with the densification \& pruning from $300$ to $2100$ iterations at an interval of $300$ steps. The prune-only phase is conducted at a scaling factor threshold of $0.008$ from $2400$ to $3300$ every $300$ steps. The overall framework is trained using Adam optimizer~\cite{adam}, with the betas of $[0.9, 0.99]$ and the learning rates of $5e-5$, $1e-3$, $1e-2$, $1.25e-2$, and $1e-2$ for the center position $\mu$, scaling factor $s$, rotation quaternion $q$, color $c$, and opacity $\alpha$, respectively.

\noindent\textbf{Training and Implementation Setups.} 
The framework is implemented in PyTorch~\cite{paszke2019pytorch} based on ThreeStudio~\cite{guo2023threestudio}. We use the camera distance range of $[1.5, 2.0]$, fovy range of $[40^\circ, 70^\circ]$, elevation range of $[-30^\circ, 30^\circ]$, and azimuth range of $[-180^\circ, 180^\circ]$. During the $1200$ to $3600$ iterations, we zoom into the head region with camera distance range of $[0.4, 0.6]$ at $25\%$ probability to enhance facial quality. The dual-branch SDS loss weights for RGB and depth $\lambda_1$, $\lambda_2$ are both set as $0.5$. We use the training resolution of $1024$ with a batch size of $8$ and the whole optimization process takes one hour on a single NVIDIA A100 (40GB) GPU.

\begin{figure*}[t]
    \centering
    \includegraphics[width=\linewidth]{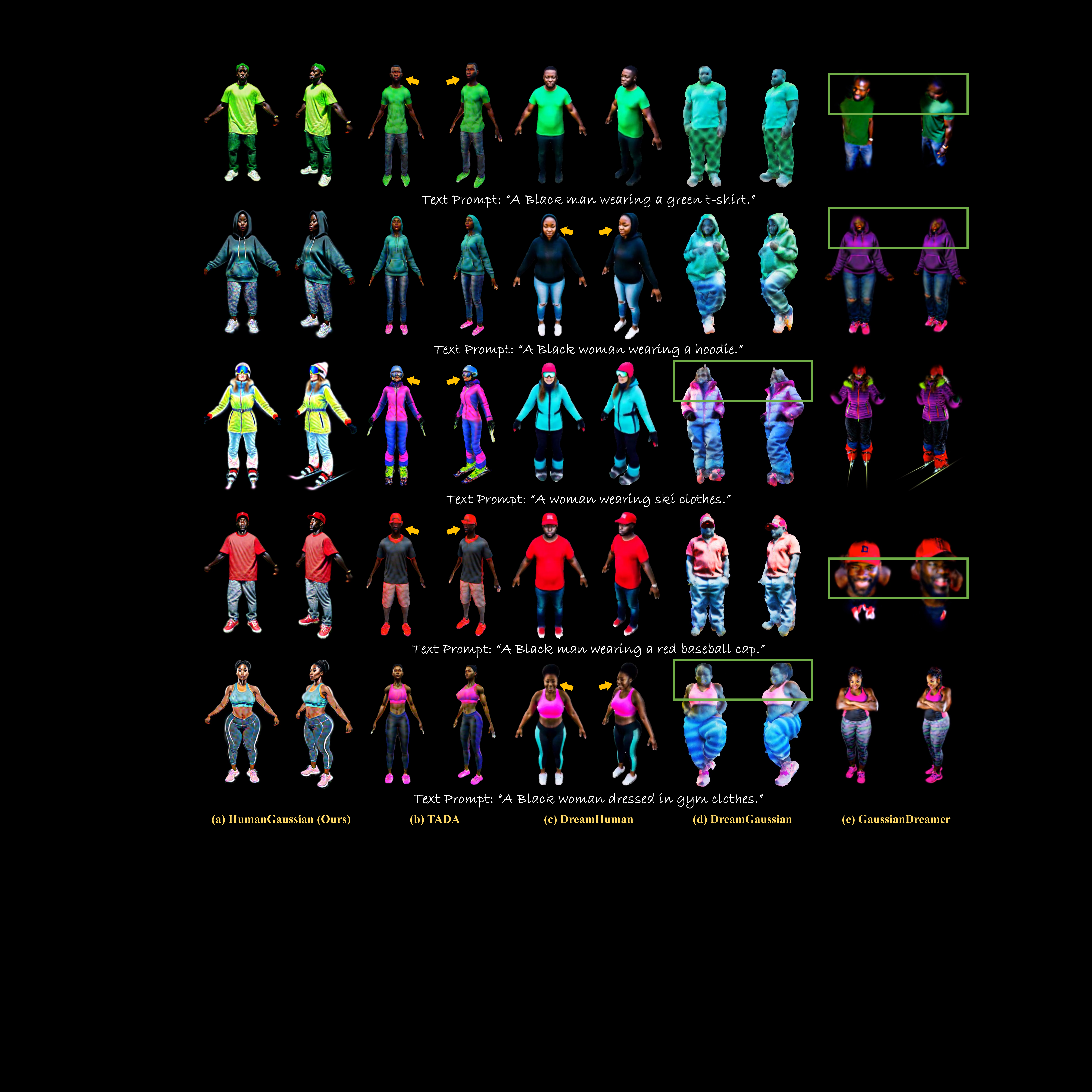}
    \vspace{-15pt}
    \caption{\textbf{Visual Comparisons with Text-to-3D and 3D Human Models.} We compare with recent state-of-the-art baselines on five different prompts, each showing two camera views. Note that the textural unrealism and blurriness are highlighted with \textbf{\textcolor[rgb]{1, 0.75, 0}{yellow}} arrows; the geometric artifacts are highlighted with \textbf{\textcolor[rgb]{0.44, 0.68, 0.28}{green}} rectangles. Please kindly \textbf{zoom in for best view} and refer to demo video for more results.}
    \vspace{-10pt}
    \label{fig:compare}
\end{figure*}

\begin{figure*}[t]
    \centering
    \vspace{-13pt}
    \includegraphics[width=\linewidth]{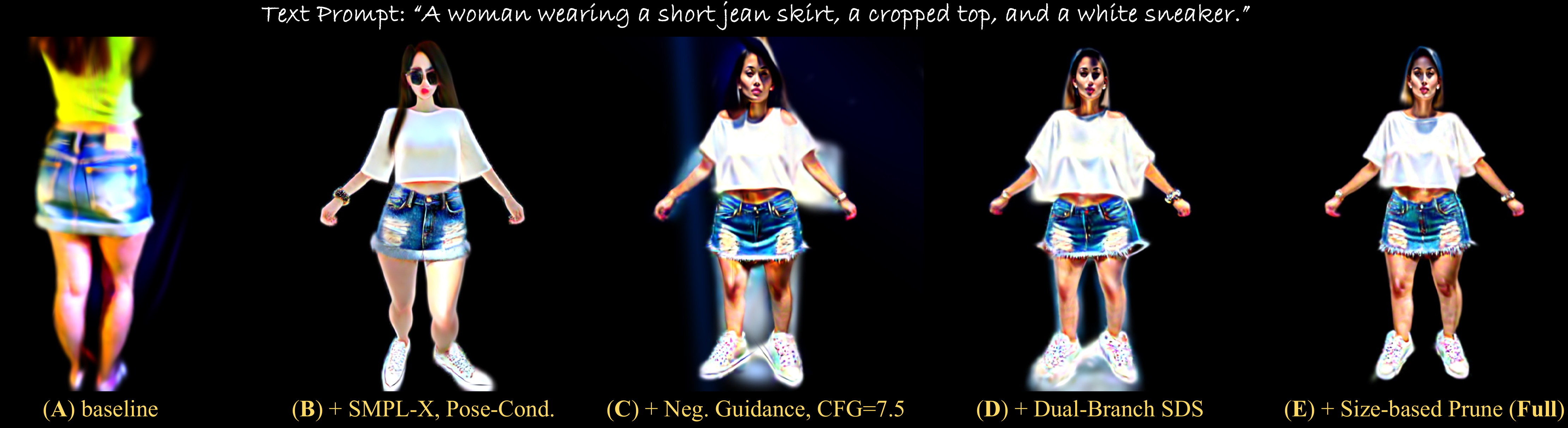}
    \vspace{-18pt}
    \caption{\textbf{Ablation Studies on HumanGaussian Module Design.} We present generation results of the human frontal view under five ablation settings for better visualization and comparisons: \textbf{(A)} \textit{baseline}; \textbf{(B)} \textit{+SMPL-X, Pose-Cond.}; \textbf{(C)} \textit{+Neg. Guidance, CFG=$7.5$}; \textbf{(D)} \textit{+Dual-Branch SDS}; \textbf{(E)} \textit{+Size-based Prune}. The detailed ablation setting designs and result analysis are elaborated in Sec.~\ref{sec:ablation}.}
    \vspace{-13pt}
    \label{fig:ablation}
\end{figure*}

\subsection{Text-Driven 3D Human Generation}
\label{sec:exp.3}
\noindent\textbf{Comparison Methods.} 
We compare with two categories of recent SOTA works: \textbf{1)} General text-to-3D methods, including DreamGaussian~\cite{tang2023dreamgaussian} and GaussianDreamer~\cite{yi2023gaussiandreamer}, which are two concurrent models that also use 3DGS as neural representation. \textbf{2)} Specialized text-to-3D human models, where we show visual comparisons with TADA~\cite{liao2023tada} and DreamHuman~\cite{kolotouros2023dreamhuman}.
Note that since the pre-trained model or training code of DreamHuman is not officially released, we directly use the results from their paper and project page.

\noindent\textbf{Qualitative Analysis.} 
The visualized generation results are shown in Fig.~\ref{fig:teaser}, where the proposed framework can generate high-quality 3D humans with realistic appearance and fine-grained geometry (\textit{e.g.}, the cloth wrinkles and accessories of ear-ring in the 4-th row). To further validate the effectiveness of our method, we show visual comparisons with recent works~\cite{liao2023tada, kolotouros2023dreamhuman, tang2023dreamgaussian, yi2023gaussiandreamer} in Fig.~\ref{fig:compare}. Note that we augment two 3DGS-based baselines with multi-view~\cite{shi2023mvdream} or Shap-E~\cite{jun2023shap} prior to make their results stronger, following the authors' instructions. It can be seen that \textbf{HumanGaussian} achieves superior performance, rendering more realistic human appearance, more coherent body structure, better view consistency, and more fine-grained detail capturing.

\noindent\textbf{User Study.}
We conduct a user study to better reflect 3D human quality. For fair comparisons, we randomly sample 30 prompts from DreamHuman~\cite{kolotouros2023dreamhuman} and involve 17 participants for subjective evaluation. The users are asked to rate three aspects: (1) \textit{Texture Quality}; (2) \textit{Geometry Quality}; (3) \textit{Text Alignment}. The rating scale is 1 to 5, with the higher the better. The results are reported in Table~\ref{table:userstudy}, where our method performs the best on all three aspects.

\subsection{Ablation Study}
\label{sec:ablation}
We present the ablation studies on two key modules of our framework in Figure~\ref{fig:ablation}. In particular, we explore the following aspects: \textbf{1)} structural prior from SMPL-X and pose condition; \textbf{2)} annealed negative prompt guidance; \textbf{3)} dual-branch SDS; and \textbf{4)} size-conditioned Gaussian pruning, by \textit{gradually adding each component} as our ablation settings: \textbf{(A)} \textit{baseline} that naively generates 3D human with 3DGS; \textbf{(B)} \textit{+SMPL-X, Pose-Cond.}, which uses SMPL-X for initialization and skeleton for SDS guidance model conditioning; \textbf{(C)} \textit{+Neg. Guidance, CFG=$7.5$} that incorporates annealed negative prompt guidance with nominal CFG scale of $7.5$; \textbf{(D)} \textit{+Dual-Branch SDS} that extends the depth-branch SDS instead of solely relying on supervisions from pixel-space; \textbf{(E)} \textit{+Size-based Prune} that removes tiny Gaussian artifacts. 

\noindent\textbf{Ablation Analysis.}
By comparing the visual quality among different settings, we can clearly see the effectiveness of each design: \textbf{1)} Different from \textbf{Config A} that suffers from incorrect articulations with multi-face Janus problem, explicit body prior and pose guidance in \textbf{Config B} guarantee coherent body structures. \textbf{2)} With the cleaner negative classifier score in \textbf{Config C}, we manage to bypass cartoonish styles or over-saturated patterns caused by large CFG scales and achieve realistic appearance. \textbf{3)} The dual-branch SDS (\textbf{Config D}) further provides joint texture-structure guidance to regularize geometric errors near limb and hair. \textbf{4)} Thanks to the size-conditioned Gaussian removal in a prune-only phase, in the full model (\textbf{Config E}) we eliminate floating artifacts near the human surface. Please kindly refer to our demo video for clearer ablation result comparisons.

\begin{table}[h]
\setlength{\tabcolsep}{3.0pt}
\small
  \centering
  \begin{tabular}{lcccccc}
    \toprule
    Methods & Text. Qual. & Geo. Qual. & Text Align. \\
    \midrule
    TADA~\cite{liao2023tada} & 3.76 & 3.53 & 4.35 \\
    DreamHuman~\cite{kolotouros2023dreamhuman} & 3.41 & 3.65 & 4.24 \\
    \midrule
    DreamGaussian~\cite{tang2023dreamgaussian} & 2.18 & 2.88 & 2.94 \\
    GaussianDreamer~\cite{yi2023gaussiandreamer} &  2.94 & 3.00 & 3.06 \\
    \midrule
    \textbf{HumanGaussian (Ours)} & \textbf{4.24} & \textbf{3.88} & \textbf{4.71} \\
    \bottomrule

  \end{tabular}
  \vspace{-10pt}
  \caption{\textbf{User Study Results.} We conduct subjective evaluations on 3D human generation quality from three aspects: (1) \textit{Texture Quality}; (2) \textit{Geometry Quality}; (3) \textit{Text Alignment}. The Mean Opinion Scores (MOS) protocol is adopted to rate based on the score scale of 1 to 5, with 1 being the poorest and 5 being the best.}
  \vspace{-2pt}
  \label{table:userstudy}
\end{table}

\begin{table}[h]\footnotesize
\setlength{\tabcolsep}{2.0pt}
  \centering
  \begin{tabular}{lccccc}
    \hline
    Metrics & TADA~\cite{liao2023tada} & DreHum.~\cite{kolotouros2023dreamhuman} & DreGau.~\cite{tang2023dreamgaussian} & GauDre.~\cite{yi2023gaussiandreamer} & \textbf{Ours} \\
    \hline
    CLIP$\uparrow$ & 30.13 & 29.98 & 24.77 & 25.12 & \textbf{30.82}\\
    \hline
    Aes.$\uparrow$ & 5.709 & 5.170 & 4.533 & 4.641 & \textbf{6.436}\\
    \hline
    HPSv2$\uparrow$ & 0.253 & 0.248 & 0.229 & 0.231 & \textbf{0.262}\\
    \hline
  \end{tabular}
  \vspace{-10pt}
  \caption{\textbf{Additional Quantitative Results.} We further evaluate T2I metrics of CLIP, aesthetic, and HPS scores on frontal views.}
  \vspace{-7pt}
\end{table}
\section{Discussion}
\label{sec:conclusion}
\noindent\textbf{Conclusion.} 
In this paper, we propose an efficient yet effective framework \textbf{HumanGaussian} for high-quality 3D human generation with fine-grained geometry and realistic appearance. We first propose \textit{Structure-Aware SDS} that simultaneously optimizes human appearance and geometry. Then we devise an \textit{Annealed Negative Prompt Guidance} to guarantee realistic results without over-saturation and eliminate floating artifacts. Extensive experiments demonstrate the superior efficiency and competitive quality of our framework, rendering vivid 3D humans under diverse scenarios.

\noindent\textbf{Limitations and Future Work.} 
As an early attempt in taming 3DGS for text-driven 3D human, our method creates realistic results. However, due to the limited performance of existing T2I models for hand and foot generation, we find it sometimes fails to render these parts faithfully. We will explore these problems in future work.

\noindent\textbf{Acknowledgement.} 
This study is supported by the Ministry of Education, Singapore, under its MOE AcRF Tier 2 (MOE-T2EP20221- 0012) and NTU NAP.


{
    \small
    \bibliographystyle{ieeenat_fullname}
    \bibliography{main}
}


\end{document}